\documentclass[10pt,twocolumn,letterpaper]{article}

\usepackage{cvpr}              % To produce the CAMERA-READY version
\usepackage[table]{xcolor}
\definecolor{cvprblue}{rgb}{0.21,0.49,0.74}
\usepackage[pagebackref,breaklinks,colorlinks,allcolors=cvprblue]{hyperref}
\usepackage{multirow}
\usepackage{makecell}  % for line breaks in headers
\usepackage{algorithm}
\usepackage{algorithmic}

\usepackage{pifont}
\newcommand{\cmark}{\ding{51}}  % ✓
\newcommand{\xmark}{\ding{55}}  % ✗

\def\paperID{42698} % *** Enter the Paper ID here
\def\confName{CVPR}
\def\confYear{2026}

\title{Mostly Text, Smart Visuals: Asymmetric Text-Visual Pruning for Large Vision-Language Models}

\author{
Sijie Li\\
University of Sheffield, UK\\
{\tt\small sli256@sheffield.ac.uk}
\and
Biao Qian\\
Tsinghua University, China\\
{\tt\small hfutqian@gmail.com}
\and
Jungong Han\thanks{Corresponding author}\\
Tsinghua University, China\\
{\tt\small jghan@tsinghua.edu.cn}
}

\begin{document}
\maketitle
\begin{abstract}
Network pruning is an effective technique for enabling lightweight Large Vision-Language Models (LVLMs), which primarily incorporates both weights and activations into the importance metric. However, existing efforts typically process calibration data from different modalities in a unified manner, overlooking modality-specific behaviors. This raises a critical challenge: how to address the divergent behaviors of textual and visual tokens for accurate pruning of LVLMs.  To this end, we systematically investigate the sensitivity of visual and textual tokens to the pruning operation by decoupling their corresponding weights, revealing that: (i) the textual pathway should be calibrated via text tokens, since it exhibits higher sensitivity than the visual pathway; (ii) the visual pathway exhibits high redundancy, permitting even 50\% sparsity. Motivated by these insights, we propose a simple yet effective \underline{A}symmetric \underline{T}ext-\underline{V}isual Weight \underline{Pruning} method for LVLMs, dubbed \textbf{ATV-Pruning}, which establishes the importance metric for accurate weight pruning by selecting the informative tokens from both textual and visual pathways. Specifically, ATV-Pruning integrates two primary innovations: first, a calibration pool is adaptively constructed by drawing on all textual tokens and a subset of visual tokens; second, we devise a layer-adaptive selection strategy to yield important visual tokens. Finally, extensive experiments across standard multimodal benchmarks verify the superiority of our ATV-Pruning over state-of-the-art methods. Code is available at \href{https://github.com/LezJ/ATV-Pruning}{https://github.com/LezJ/ATV-Pruning}.
\end{abstract}

\vspace{-1em}    
\section{Introduction}
\label{sec:intro}

Large Vision-Language Models (LVLMs) have exhibited outstanding capabilities in various tasks, such as visual question answering \cite{alayrac2022flamingo, liu2023visual, liu2024llavanext}, embodied intelligence \cite{driess2023palm, zitkovich2023rt}, and video understanding \cite{zhang2023video, cheng2024videollama}. Despite the remarkable performance, LVLMs are typically parameter-heavy and computationally demanding, which pose critical challenges for deployment on resource-constrained devices, \emph{e.g.}, robots and autonomous vehicles. 
\begin{figure}[t]
  \centering
  \includegraphics[width=\linewidth]{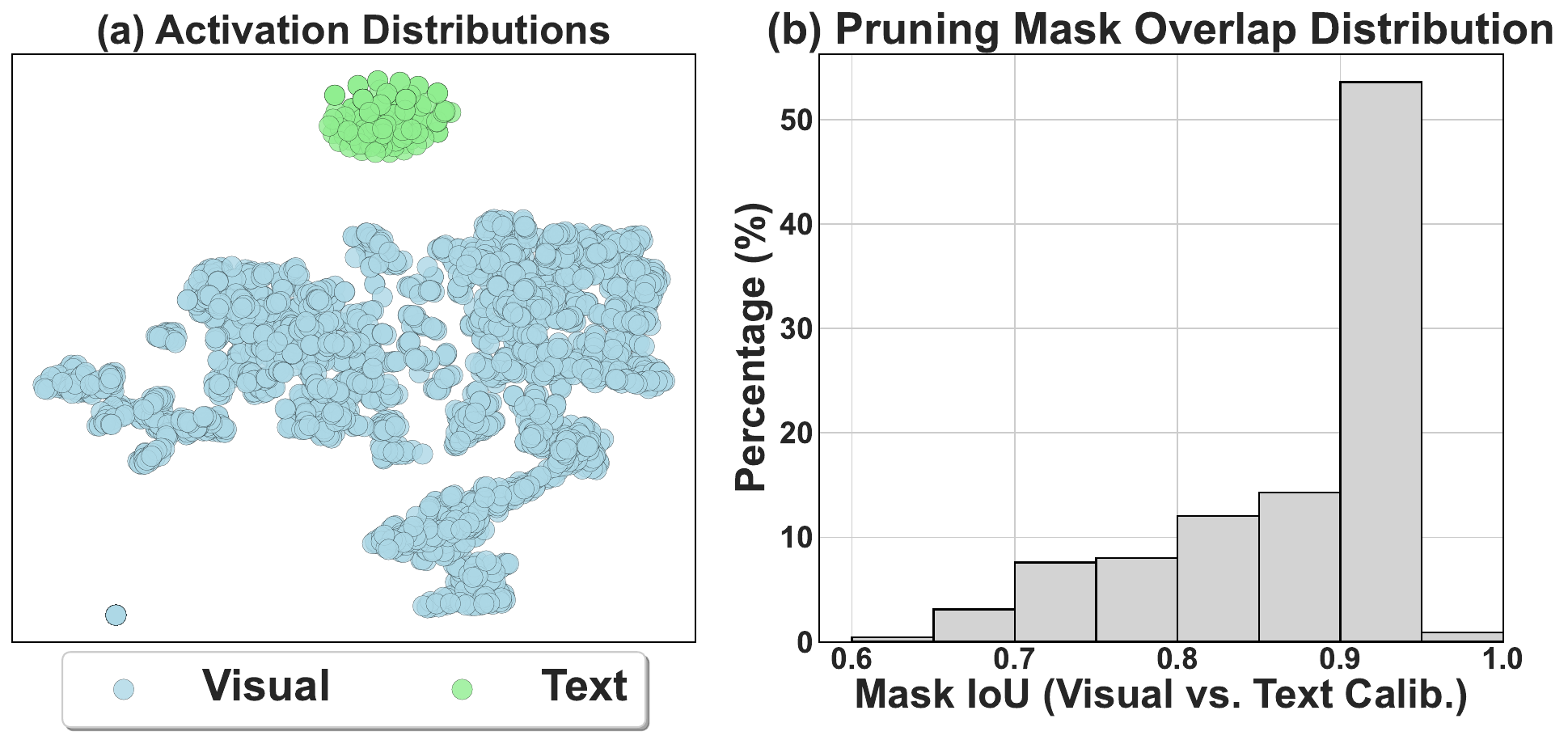}
  \caption{
    \textbf{Illustration of the divergent statistical characteristics across different modalities}, manifesting as: (a) \emph{activation representation}: the textual and visual activations occupy distinct clustered regions in the representation space (t-SNE visualization); and (b) \textit{pruning importance}: the pruning
masks derived from the text-only and visual-only calibration data exhibit a broad IoU distribution (taking 50\% sparsity level as an example)
  }
  \label{fig:figure1}
  \vspace{-1.5em}
\end{figure}

To remedy this issue, weight pruning has emerged as a powerful technique to remove redundant model parameters. In the unimodal Large Language Models (LLMs) domain, post-training methods like SparseGPT~\cite{frantar2023sparsegpt} and efficient activation-aware approaches~\cite{sunsimple, das2023beyond, yang2025wanda++} (e.g., Wanda~\cite{sunsimple}) have achieved remarkable success with a small calibration set, i.e., a held-out set of representative samples used to estimate pruning criteria. Nevertheless, naively applying these LLMs-centric techniques to LVLMs generally suffers from suboptimal performance, since different input modalities can exhibit distinct sensitivity to the pruned sub-networks, harming cross-modal alignment and semantic integrity. Recently, several LVLM-specific studies have proposed to seek optimal block-wise sparsity allocations~\cite{sungecoflap,liang2025efficientllava} and explore modality-aware schemes~\cite{lee-etal-2025-tamp}. For instance, TAMP~\cite{lee-etal-2025-tamp} uses multimodal statistics to guide calibration token selection, but still treats visual and textual tokens within a shared selection process. These approaches tend to leverage mixed-modal activations in a largely uniform manner for importance assessment, hence they still fail to address the persistent issue of modality heterogeneity.

We identify the divergent statistical characteristics across different modalities as two aspects: \emph{activation representation}, where the textual and visual activations occupy distinct clustered regions in the representation space (see Fig.~\ref{fig:figure1}(a)); and \emph{pruning importance}, where the pruning masks derived from the text-only and visual-only calibration data exhibit a broad IoU distribution (see Fig.~\ref{fig:figure1}(b)), indicating that the weights deemed important by the textual modality can differ significantly from those identified by the visual modality. The above motivates us to delve into \emph{how different modalities behave differently under pruning operations} and \emph{how to effectively perform activation-aware pruning in LVLMs}.

To answer these questions, we conduct a series of motivation experiments that decouple text-associated and visual-associated parameters within Transformer blocks and independently study the sensitivity of different input modalities.
Our motivation experiments reveal the following findings:

\noindent \emph{(i) The textual pathway is highly sensitive to the selection of the calibration pool.} 
Pruning the textual pathway with visual-only or mixed calibration pool substantially degrades the accuracy. For instance, on SQA$_\text{img}$, the model performance collapses to 35.85 and 11.1 at 60\% sparsity level with mixed or visual-only calibration pool; in contrast, text-only calibration pool reaches 61.58 (see Tab.~\ref{tab:sensitivity}).

\noindent \emph{(ii) The visual pathway is highly redundant and robust.} Pruning the visual pathway to 60\% sparsity preserves over 99.25\% performance across all calibration pools, \emph{even with text-only calibration} (Tab.~\ref{tab:sensitivity}), suggesting that text calibration already identifies most visually crucial weights, with only a few visual tokens needed to complement the text pool and capture the remaining visual-specific parameters.

Motivated by these observations, in this paper, we propose a novel Asymmetric Text-Visual Weight Pruning method for LVLMs, named \textbf{ATV-Pruning}, which explicitly accounts for the inherent differences between textual and visual modalities during pruning. 
Specifically, ATV-Pruning comprises two key components:
(i) a cross-modal calibration pool is constructed by adaptively integrating all textual tokens with a sparse subset of visual tokens; to achieve this, (ii) a block-adaptive selection strategy is devised to identify and retain the most crucial visual tokens.
The resulting pool is then used to evaluate the weight importance scores for weight pruning of LVLMs. Extensive experiments across standard multimodal benchmarks and models demonstrate the merits of ATV-Pruning over state-of-the-art methods.

Our major contributions are summarized as follows:
\begin{itemize}
    \item We provide novel insights into the asymmetric pruning sensitivity across modalities: (i) text-side pruning is highly sensitive and must be calibrated with text tokens, while (ii) the visual pathway is highly robust, admitting 60\% unstructured sparsity with minimal degradation.
    \item We propose ATV-Pruning, a simple yet effective pruning framework that adaptively constructs a cross-modal calibration pool by integrating full textual tokens with a compact, layer-adaptive subset of visual tokens. 
    \item Comprehensive evaluation on nine standard multimodal benchmarks demonstrate the superiority of ATV-Pruning over state-of-the-art baselines. 
\end{itemize}

\section{Related Work}
\label{sec:related}

\paragraph{LVLMs and efficiency.}
The remarkable success of Large Language Models (LLMs)~\cite{ouyang2022training, touvron2023llama, bai2023qwen} in text-based reasoning has spurred the development of Large Vision-Language Models (LVLMs)~\cite{alayrac2022flamingo, liu2023visual, liu2024llavanext}, which extend these capabilities into the visual domain. Early LVLMs demonstrated strong performance on foundational tasks like visual question answering \cite{alayrac2022flamingo, liu2023visual}. 
Recently, the field has rapidly advanced towards more complex and data-intensive capabilities. For instance, LLaVA-NeXT~\cite{liu2024llavanext} pushes performance in fine-grained reasoning and OCR by processing images at significantly higher resolutions. Simultaneously, models like LLaVA-OneVision~\cite{li2024llava} and Qwen2-VL~\cite{wang2024qwen2} have been developed as unified systems that can process single-image, multi-image, and video inputs to perform more complex reasoning.
While impressive, these advancements directly exacerbate the computational challenges. They are expensive to serve in practice: the LLM backbone remains large, and supporting high-resolution images or long video sequences introduces exceptionally long visual token sequences. Consequently, the inference latency and memory cost scale accordingly, making deployment on resource-constrained systems a critical bottleneck.

Several directions have been explored to improve efficiency.
One line aims to shrink model size directly via smaller backbones~\cite{chu2023mobilevlm, zhou2024tinyllava}.
Another line applies quantization~\cite{frantar2022gptq, wang2024q}, reducing precision of weights and/or activations for faster inference.
A third line targets sequence redundancy through visual token pruning~\cite{chen2024image,alvar2025divprune,zhao2025stitch}, which drops a large fraction of vision tokens with minimal loss.
Our work is orthogonal: we focus on \emph{weight pruning} to sparsify the shared backbone and improve efficiency.
We additionally show that redundancy is not only in visual tokens but also in the visual-processing pathway parameters, which can tolerate surprisingly high sparsity.

\vspace{-1.3em}

\noindent \paragraph{Weight pruning methods.}
For LLMs, SparseGPT~\cite{frantar2023sparsegpt} prunes weights by approximately minimizing output error using a local Hessian-based update.
This yields strong performance but requires an iterative weight update procedure per layer, which adds pruning overhead.
Wanda~\cite{sunsimple} instead proposes an activation-aware importance score based on weight magnitude and input activation norms; it achieves competitive retention without expensive weight updates, making it attractive for fast post-hoc sparsification.

Extending pruning to LVLMs, prior work has primarily examined how to allocate sparsity budgets across Transformer blocks to account for the multimodal nature~\cite{sungecoflap,liang2025efficientllava,lee-etal-2025-tamp}, whereas our work focuses on improving the pruning criterion itself. 
Among these methods, TAMP~\cite{lee-etal-2025-tamp} further introduces Adaptive Multimodal Input Activation: it selects a subset of highly attended, diverse text and visual tokens as calibration input for Wanda scoring. While this improves performance over vanilla Wanda, the selection remains heuristic and introduces significant extra compute. Crucially, the modality-specific pruning behavior is not explicitly examined. In contrast, we start from a modality-disentangled analysis that reveals the distinct sensitivities of text and visual pathways under different calibration pools; guided by these findings, ATV-Pruning applies an asymmetric token-selection strategy—keeping all text tokens while layer-adaptively sampling a small set of informative visual tokens—to obtain more stable pruning.

\section{Methodology}
\label{sec:method}

Our goal is to prune the LLM backbone of an LVLM under high unstructured sparsity, while preserving both linguistic capability and multimodal reasoning quality.
We argue that, in multimodal settings, \emph{the choice of calibration tokens used to estimate activation statistics is critical}, as activation distributions differ substantially across modalities.
In this section, we begin by revisiting the activation-aware pruning scheme (Sec.~\ref{sec:pre}), represented by Wanda ~\cite{sunsimple}. Following that, we perform the motivation experiments to offer insights into the weight pruning for LVLMs (Sec.~\ref{sec:modality_analysis}). Building on this, we elaborate our proposed Asymmetric Text-Visual Weight Pruning method (Sec.~\ref{sec:atv_pruning}).

\subsection{Preliminaries}
\label{sec:pre}
Our pruning recipe is built on top of \emph{Wanda}~\cite{sunsimple}, a post-hoc, activation-aware method, which evaluates the weight importance scores by combining weight magnitude and estimated input activation.

\noindent \textbf{Weight importance scoring.}
Consider a linear layer with weight matrix $\mathbf{W} \in \mathbb{R}^{d_\text{out} \times d_\text{in}}$ and corresponding input activations $\mathbf{X} \in \mathbb{R}^{N \times d_\text{in}}$, where $N$ indexes token positions across a calibration batch.
An importance score is assigned to each scalar weight $\mathbf{W}_{ij}$ via
\begin{equation}
\label{eq:wanda_score}
\mathbf{I}_{ij}
\;=\;
\big| \mathbf{W}_{ij} \big| \cdot
\big\| \mathbf{X}_{j} \big\|_{2},
\end{equation}
where $\mathbf{X}_{j} \in \mathbb{R}^{N}$ denotes the $j$-th input channel across all tokens in the calibration batch, and $\|\cdot\|_2$ is the Euclidean norm.
Intuitively, a weight is considered important when it is \emph{both large in magnitude and frequently activated}.
With $\mathbf{I}_{ij}$ for all entries in $\mathbf{W}$, we can prune the lowest-scoring $\rho\%$ weights, yielding unstructured sparsity of the layer.

\noindent \textbf{Standard calibration for LLMs.}
In unimodal LLM pruning, $\|\mathbf{X}_{j}\|_{2}$ in Eq.~\ref{eq:wanda_score} is computed over \emph{all} tokens from a small text-only calibration corpus.
This implicitly assumes:
(i) all inference-time tokens ``behave similarly enough'' to share statistics,
and
(ii) the model is dominated by a single modality, so the calibration pool is homogeneous.

\noindent \textbf{Modality-Agnostic Calibration Pool for LVLMs}. 
Unlike LLMs, a typical input sequence for LVLMs interleaves text tokens with visual tokens. When the weight pruning shifts from LLMs to LVLMs, the naive strategy is to apply the principles of Wanda to LVLMs, which reuses Eq.~\ref{eq:wanda_score} and compute $\|\mathbf{X}_{j}\|_{2}$ over a calibration pool that comprises all multimodal tokens (i.e., both text and visual tokens).
We refer to this ``use everything'' approach as the \emph{modality-agnostic calibration pool}.
It is straightforward to implement and remains a common practical choice in LVLM pruning pipelines.
Although several recent studies have begun to consider the multimodal nature of calibration data~\cite{lee-etal-2025-tamp,liang2025efficientllava, sungecoflap}, a systematic investigation of how different calibration sources (text-only, image-only, or mixed) influence pruning behavior across text- and image-associated weights has been lacking.
To bridge this gap, we conduct a controlled, \emph{modality-aware} sensitivity analysis as discussed in the next.

\subsection{Motivation Investigation: Modality-Aware Sensitivity Analysis}
\label{sec:modality_analysis}

\begin{figure}[t]
    \centering
    \includegraphics[width=0.8\linewidth]{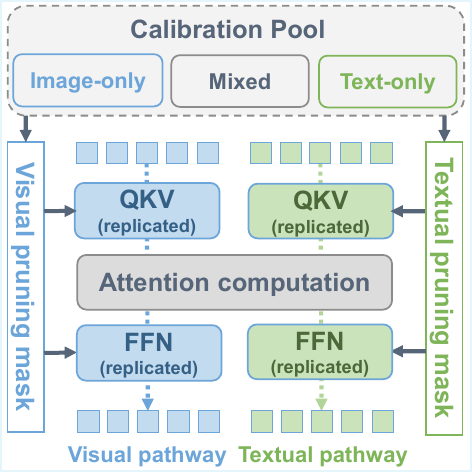}
    \caption{\textbf{Modality decoupling via MoT probe.} For each Transformer block, the QKV and FFN layers are \textbf{replicated into visual and textual pathways}, which process their respective token types. Independent pruning masks are derived for each pathway using activation statistics from \textit{text-only}, \textit{image-only}, or \textit{mixed} calibration pools.
    This setup enables controlled comparison of modality-specific pruning sensitivity.}
    \label{fig:decoupling}
    \vspace{-1.3em}
\end{figure}

To provide insights into the effects of textual and visual modalities on the weight pruning process, we aim to delve into the following critical question: \emph{for the activation-aware pruning scheme in LVLMs, which type of tokens should be exploited to estimate activation statistics while preserving overall performance?} Therefore, we design a series of motivation experiments, where the textual and visual pathways are explicitly disentangled without harming the original computation flows, with LLaVA-NeXT~\cite{liu2024llavanext} (more experimental details in the Supplementary). Specifically, we introduce a Mixture-of-Transformer (MoT)~\cite{shi2024lmfusion, deng2025emerging} analysis probe (Fig.~\ref{fig:decoupling}), where (i) the shared weights for each transformer block are replicated and allocated to the textual and visual pathway; (ii) the textual and visual pathway are independently pruned under varied calibration pools. In particular, we select either of the textual, visual or both type tokens to compute $\|\mathbf{X}_{j}\|_{2}$, which is further utilized to evaluate the weight importance scores $I_{ij}$ in Eq.~\ref{eq:wanda_score}. According to the type of selected tokens, the calibration pool involves the following three cases: \textbf{text-only calibration pool} to use only positions corresponding to text tokens; \textbf{image-only calibration pool} to use only positions corresponding to visual tokens;  \textbf{mixed calibration pool} to use all tokens from the full multimodal sequence.
With the above settings, Tab.~\ref{tab:sensitivity} reports experimental results under the $2{\times}3$ design at $50\%$ and $60\%$ sparsity, and reveals the following critical findings.

\renewcommand{\multirowsetup}{\centering}
\definecolor{mygray}{gray}{.92}
\definecolor{mygreen2}{RGB}{238, 243, 243}
\definecolor{ForestGreen}{RGB}{34,139,34}

\begin{table}[t]
    \centering
    \footnotesize
    \setlength{\tabcolsep}{3pt}
    \renewcommand{\arraystretch}{1.2}
    \resizebox{0.95\linewidth}{!}{
    \begin{tabular}{l l | *{3}{>{\centering\arraybackslash}p{1.05cm}} | >{\centering\arraybackslash}p{1.15cm}}
        \textbf{Pathway} & \textbf{Calibration} 
        & \textbf{MMB} & \textbf{SQA$_\text{img}$} & \textbf{VizWiz} 
        & \makecell[c]{\textbf{Average}} \\
        \midrule

        \multicolumn{2}{c|}{\textcolor{gray}{Dense}} & \textcolor{gray}{72.16} & \textcolor{gray}{73.28} & \textcolor{gray}{57.65} &  \textcolor{gray}{100\%} \\
        \midrule
        
        \rowcolor{mygray}
        \multicolumn{6}{c}{\textit{50\% Sparsity}} \\
        \multirow{3}{*}{Text} 
            & Text
               & 65.03  & 70.55 & 62.94 & 98.26\% \\
            & Visual          
                & 61.77 & 67.67 & 60.56 & 94.33\% \\
            & Mixed          
                & 64.00 & 69.01 & 60.36 & 95.86\% \\
        \midrule
        \multirow{3}{*}{Visual} 
            & Text
               & 71.39  & 74.17 & 58.03 & 100.27\% \\
            & Visual          
                & 70.87 & 73.13 & 57.71 & 99.37\% \\
            & Mixed          
                & 71.56 & 73.92 & 57.87 & 100.14\% \\
        \midrule
        
        \rowcolor{mygray}
        \multicolumn{6}{c}{\textit{60\% Sparsity}} \\
        \multirow{3}{*}{Text} 
            & Text
               & 52.15  & 61.58 & 56.30 & 84.65\% \\
            & Visual          
                & 41.58 & 11.11 & 46.11 & 50.92\% \\
            & Mixed          
                & 44.76 & 35.85 & 48.41 & 64.97\% \\
        \midrule
        \multirow{3}{*}{Visual} 
            & Text
               & 71.05  & 73.67 & 58.32 & 100.05\% \\
            & Visual          
                & 70.53 & 73.08 & 57.81 & 99.25\% \\
            & Mixed          
                & 70.88 & 73.38 & 57.85 & 99.57\% \\

    \end{tabular}
    }
    \caption{\textbf{Sensitivity analysis of both modality pathways} across different calibration token sources, benchmarks, and sparsity.}
    \label{tab:sensitivity}
    \vspace{-2em}
\end{table}

\begin{itemize}
\item \textbf{\emph{Finding A: Asymmetry calibration sensitivity across textual and visual pathways.}}
For the \emph{textual pathway} pruning, the best retention is consistently achieved when $\|\mathbf{X}_j\|_2$ in Eq.~\ref{eq:wanda_score} is computed using \emph{text-only} calibration pool. Meanwhile, using mixed calibration pool behaves better than image-only calibration pool.
In contrast, for \emph{visual pathway} pruning, using different calibration pools receives the similar pruning performance, hence the visual pathway is less sensitive to the type of calibration pool compared to the textual pathway.
Such fact indicates that text-anchored calibration is necessary to safeguard linguistic and reasoning capability, while using text-driven calibration generally works well even when pruning weights processing visual tokens.

\item \textbf{\emph{Finding B: The visual pathway possesses significant weight-level redundancy.}}
We observe that, when applying $50\%$ or $60\%$ unstructured sparsity solely to the visual pathway, the pruning performance is comparable to the dense baseline, pointing to substantial over-parameterization in visual processing. Combined with \emph{Finding A}, the facts suggest that, during calibration, it is neither necessary nor beneficial to treat all visual tokens as equally informative. Instead, a compact subset of salient visual evidence is sufficient to protect vision performance. Note that this redundancy differs from token-level redundancy explored by visual token pruning studies~\cite{chen2024image, alvar2025divprune, zhao2025stitch}, where the former lies in parameters within the backbone, the latter in input token sequence.
\end{itemize}

Interestingly, on VizWiz at $50\%$ sparsity, we observe a performance gain over the dense model—an effect also noted in prior LVLM pruning study~\cite{liang2025efficientllava}, suggesting that removing ambiguous neurons can improve performance on certain benchmarks.
Our analysis extends this finding by showing that such beneficial pruning mainly occurs within the textual pathway, especially when applying text-only calibration.
This further reinforces that text calibration is essential, and sometimes even beneficial for LVLM pruning.
% for preserving—and sometimes even enhancing—overall performance.

\noindent\textbf{Takeaway.}
A modality-agnostic calibration pool dilutes the linguistic signal essential for protecting text-associated weights.
Calibration should be text-anchored and augmented only with a compact set of salient visual tokens.

% -------------------------
\subsection{Asymmetric Text-Visual Pruning}
\label{sec:atv_pruning}
Motivated by the findings in Sec.~\ref{sec:modality_analysis}, we propose an Asymmetric Text-Visual Weight Pruning (ATV-Pruning) method for LVLMs, which performs modality-aware weight pruning by considering the sensitivity difference of textual and visual pathways towards calibration source. We first discuss how to construct a modality-aware calibration set.

\subsubsection{Modality-Aware Calibration Set}
The above investigation conveys two principles for constructing an effective calibration set: textual tokens are essential for preserving language competency; and a subset of visual tokens is sufficient to maintain robust vision ability. 
Building on that, we replace the conventional “use everything’’ calibration pool with a \emph{modality-aware calibration set} that
(i) includes \emph{all text tokens} and
(ii) selecting a \emph{small, layer-adaptive subset of high-impact visual tokens}.
Formally, for each calibration sample, let $\mathcal{T}$ be the set of all text-token positions, and let $\mathcal{V}_{\text{sub}}$ be a block-adaptive subset of salient visual-token positions.
For each transformer block, the modality-aware calibration set is defined as 
\begin{equation}
\label{eq:scal_def}
\mathcal{S}_{\mathrm{cal}}
\;=\;
\mathcal{T}
\,\cup\,
\mathcal{V}_{\text{sub}}
\end{equation}
This set is then used to evaluate the importance score according to Eq.~\ref{eq:wanda_score}. Next, we discuss the critical question of how to perform block-adaptive visual selection.

\subsubsection{Block-Adaptive Visual Selection}
\label{sec:adaptive_visual_selection}

\noindent\textbf{Visual token saliency in pruning.}
We posit that visual tokens do not contribute equally to optimal calibration statistics, and their computational role varies significantly across transformer blocks. Motivated by this, we present a block-adaptive visual selection strategy guided by \emph{token saliency}.
Specifically, given a Transformer block $b$ and a visual token $v$ from the calibration set, we define its saliency score as
\begin{equation}
\label{eq:saliency_def}
s_v \;=\; \phi_b(v),
\end{equation}
where $\phi_b(\cdot)$ is a block-specific scoring function that evaluates the importance of $v$ at block $b$.

\noindent \textbf{Instantiating saliency via visual drift.}
While various choices of $\phi_b$ are possible, we find that a token’s \emph{representation drift} within a block is a robust proxy for its saliency. Intuitively, if a block substantially updates a visual token, that token is actively involved in the computation and should be prioritized during calibration.
Let $\mathbf{X}_{\mathrm{in},v}$ and $\mathbf{X}_{\mathrm{out},v}$ denote the input and output representations of token $v$ at block $b$. We instantiate $\phi_b$ as the cosine distance between these representations:
\begin{equation}
\label{eq:cosdist}
s_v
\;=\;
1
-
\cos\!\big(
\mathbf{X}_{\mathrm{in},v},
\,
\mathbf{X}_{\mathrm{out},v}
\big),
\end{equation}
where larger $s_v$ indicates stronger visual drift and thus higher saliency.

\begin{figure}[t]
    \centering
    \includegraphics[width=0.8\linewidth]{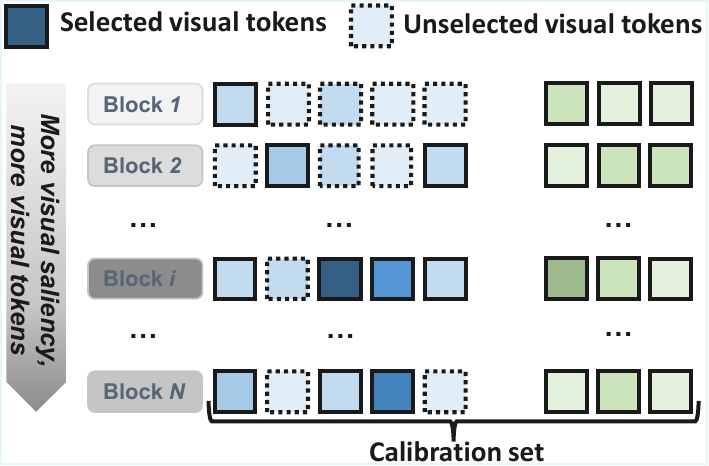}
    \caption{\textbf{Overview of ATV-Pruning;} Color intensity reflects the degree of visual saliency. Blocks with higher visual saliency keep more salient visual tokens, with all text tokens.}
    \label{fig:method_overview}
    \vspace{-1em}
\end{figure}

\noindent\textbf{From saliency to block-wise budgets.}
To capture how visually active a block is on average, we aggregate token-level saliency scores over all visual tokens observed at that block across the calibration set:
\begin{equation}
\label{eq:mean_saliency}
\bar{s}
\;=\;
\frac{1}{|\mathcal{V}_{\mathrm{all}}|}
\sum_{v \in \mathcal{V}_{\mathrm{all}}}
s_v,
\end{equation}
where $\mathcal{V}_{\mathrm{all}}$ denotes the multiset of all visual tokens at that block across the calibration batch. Empirically, $\bar{s}$ is small in early layers and larger in mid-to-late layers, reflecting the progressive strengthening of visual information processing.

\noindent We then allocate a visual-token budget $K$ to each block as
\begin{equation}
\label{eq:k_rule_saliency}
K
\;=\;
\Big\lfloor
\alpha \cdot \bar{s} \cdot n_{\text{text}}
\Big\rfloor,
\end{equation}
where $n_{\text{text}}$ is the number of text tokens for the current calibration sample and $\alpha > 0$ is a global scaling hyperparameter. Blocks with higher average visual saliency (larger $\bar{s}$) are thus allowed to retain more visual tokens, leading to a \emph{block-adaptive} allocation of visual-token budgets.

\noindent\textbf{Visual token selection given budgets.}
Given the per-block budget $K$, we perform saliency-based token selection within each calibration sample. For each sample, let $\mathcal{V}$ be the set of its visual-token indices at block $b$ and $\{ s_v : v \in \mathcal{V} \}$ the corresponding saliency scores. We keep only the top-$K$ most salient visual tokens:
\begin{equation}
\label{eq:topk_rule_saliency}
\mathcal{V}_{\text{sub}}
\;=\;
\mathrm{TopK}_{v}
\big(
\{ s_v \},
\, K
\big).
\end{equation}
We then form the modality-aware calibration set for that block via $\mathcal{S}_{\mathrm{cal}} = \mathcal{T} \cup \mathcal{V}_{\text{sub}}$ (Eq.~\ref{eq:scal_def}), and use only the positions in $\mathcal{S}_{\mathrm{cal}}$ to estimate activation norms $\|\mathbf{X}_j\|_2$ in Eq.~\ref{eq:wanda_score}.

\begin{algorithm}[t]
\caption{\textbf{ATV-Pruning procedure}}
\label{alg:atv}
\begin{algorithmic}[1]
\REQUIRE Model weights; calibration set $\mathcal{D}_{\mathrm{cal}}$; scaling factor $\alpha$; target sparsity $\rho$.
\FOR{each Transformer block}
    \STATE $\mathcal{V}_{\mathrm{all}} \leftarrow$ all visual tokens in $\mathcal{D}_{\mathrm{cal}}$ at this block
    \STATE Compute $s_v$ for all $v \in \mathcal{V}_{\mathrm{all}}$ using Eq.~\ref{eq:cosdist}
    \STATE Compute $\bar{s}$ with $s_v$ via Eq.~\ref{eq:mean_saliency}
    \FOR{each calibration sample}
        \STATE $\mathcal{T} \leftarrow$ text-token indices in the sample
        \STATE $n_{\text{text}} \leftarrow |\mathcal{T}|$
        \STATE $\mathcal{V} \leftarrow$ visual-token indices in the sample
        \STATE Compute $K = \lfloor \alpha \cdot \bar{d} \cdot n_{\text{text}} \rfloor$  (Eq.~\ref{eq:k_rule_saliency})
        \STATE Select $\mathcal{V}_{\text{sub}}$ from $\mathcal{V}$ using Eq.~\ref{eq:topk_rule_saliency}
        \STATE Form $\mathcal{S}_{\mathrm{cal}} = \mathcal{T} \cup \mathcal{V}_{\text{sub}}$  (Eq.~\ref{eq:scal_def})
    \ENDFOR
    \STATE Pool all $\mathcal{S}_{\mathrm{cal}}$ across calibration samples
    \STATE Estimate $\|\mathbf{X}_{j}\|_{2}$ only from pooled $\mathcal{S}_{\mathrm{cal}}$ positions
    \STATE Compute $\mathbf{I}_{ij} = |\mathbf{W}_{ij}| \cdot \|\mathbf{X}_{j}\|_{2}$  (Eq.~\ref{eq:wanda_score})
    \STATE Prune the lowest $\rho\%$ of weights according to $\mathbf{I}_{ij}$
\ENDFOR
\ENSURE Pruned model $\{\mathbf{W}_{\mathrm{pruned}}\}$.
\end{algorithmic}
\end{algorithm}

By construction, the resulting importance scores are \emph{text-anchored} and \emph{visually selective}, directing pruning towards weights that matter most for LVLMs' capabilities.

\noindent\textbf{Alternative saliency signals.}
Although visual drift, \emph{i.e.}, cosine distance change (Eq.~\ref{eq:cosdist}), is our default saliency signal, we also experimented with alternative choices inspired by visual token pruning studies, such as attention received from text queries~\cite{chen2024image, zhao2025stitch} and diversity-oriented token selection~\cite{alvar2025divprune}. These alternatives can induce similar block-adaptive behavior, but visual drift yields the most stable performance (see the validation experiments in Sec.\ref{ablation_selection_strategy}).

\subsubsection{Optimization Pipeline}
The above specifies our asymmetric text-visual weight pruning framework, as illustrated in Fig.~\ref{fig:method_overview}. Specifically, for each Transformer block, we kick oﬀ ATV-Pruning by performing a forward pass with the calibration set to compute the per-token saliency $s_v$ and the block-average $\bar{s}$. The budget $K$ is then derived through the global scaling factor $\alpha$, and the modality-aware calibration set $\mathcal{S}_{\mathrm{cal}}$ is formed by combining all text tokens with the top-$K$ silent visual tokens for later activation-aware pruning. This pipeline ensures that the pruning metric is dominated by linguistic signals while retaining critical visual context.
The whole pruning procedure is summarized in Algo.~\ref{alg:atv}

\section{Experiments}

\renewcommand{\multirowsetup}{\centering}
\definecolor{mygray}{gray}{.92}
\definecolor{mygreen1}{RGB}{253, 244, 244}%255, 234, 234
\definecolor{mygreen2}{RGB}{238, 243, 243}%223, 240, 240
\definecolor{ForestGreen}{RGB}{34,139,34}
\newcommand{\fg}[1]{\mathbf{\mathcolor{ForestGreen}{#1}}}
\definecolor{Forestred}{RGB}{220,50,50}
\newcommand{\fr}[1]{\mathbf{\mathcolor{Forestred}{#1}}}

\begin{table*}[t]
    \vspace{-0.75em}
    \centering
    \setlength{\tabcolsep}{3pt}
    \footnotesize
    \vspace{0.25em}
    \resizebox{\linewidth}{!}{
    \begin{tabular}{l | *{9}{>{\centering\arraybackslash}p{1.13cm}} |>{\centering\arraybackslash}p{1.15cm}}
        % \toprule[1.25pt]
        \textbf{\;Methods} & \textbf{GQA} & \textbf{MMB} & \textbf{MME} & \textbf{MMMU} & \textbf{OKVQA}  & \textbf{POPE} & \textbf{SQA}$_{\text{img}}$ & \textbf{TextVQA} & \textbf{VizWiz}  & \makecell[c]{\textbf{Average}}\\
        \midrule
        
        \textcolor{gray}{LLaVA-NeXT 8B, Dense} & \textcolor{gray}{65.34} & \textcolor{gray}{72.16} & \textcolor{gray}{1965.12} & \textcolor{gray}{40.11} & \textcolor{gray}{60.13} & \textcolor{gray}{87.84} & \textcolor{gray}{73.28} & \textcolor{gray}{65.42} & \textcolor{gray}{57.65} & \multirow{1}*{\textcolor{gray}{100\%}} \\
        \midrule

        \rowcolor{mygray}
        LLaVA-NeXT 8B & \multicolumn{10}{c}{\textit{50\% Sparsity}}\\
        SparseGPT~\cite{frantar2023sparsegpt} & 62.86 &  \textcolor{ForestGreen}{\textbf{65.38}} & 1742.80 & 36.00 & 38.17 &  \textcolor{ForestGreen}{\textbf{89.23}} & 68.12 &  \textcolor{ForestGreen}{\textbf{64.21}} & 60.12 & \multirow{1}*{91.74\%} \\
        Wanda~\cite{sunsimple}  & 62.61 & 63.83 & 1620.02 & 35.56 & 25.38 & 88.02 & 69.06 & 64.04 & 60.73  & \multirow{1}*{88.36\%} \\
        TAMP~\cite{lee-etal-2025-tamp}  & 63.15 & 64.86 & 1732.54 & 34.78 & 43.59 & 88.04 &  \textcolor{ForestGreen}{\textbf{70.00}} & 62.78 & 62.50  & 92.67\% \\
        
        \rowcolor{mygreen2}
        ATV-Pruning \scriptsize{(Ours)} & \textcolor{ForestGreen}{\textbf{63.28}} & 65.29 & \textcolor{ForestGreen}{\textbf{1801.51}} & \textcolor{ForestGreen}{\textbf{36.11}} & \textcolor{ForestGreen}{\textbf{45.34}} & 88.22 & 69.86 & 62.76 & \textcolor{ForestGreen}{\textbf{63.34}}  & \textcolor{ForestGreen}{\textbf{94.00\%}} \\
        \midrule

        \rowcolor{mygray}
        LLaVA-NeXT 8B & \multicolumn{10}{c}{\textit{60\% Sparsity}}\\
        SparseGPT~\cite{frantar2023sparsegpt} & \textcolor{ForestGreen}{\textbf{58.52}} & \textcolor{ForestGreen}{\textbf{50.60}} & \textcolor{ForestGreen}{\textbf{1586.34}} & 29.33 & 5.56 & \textcolor{ForestGreen}{\textbf{88.58}} & 55.97 & \textcolor{ForestGreen}{\textbf{61.06}} & 53.50 & \multirow{1}*{76.24\%} \\
        Wanda~\cite{sunsimple}  & 49.88 & 43.38 & 1107.72 & 27.78 & 0.90 & 82.68 & 38.67 & 54.37 & 49.85  & \multirow{1}*{64.45\%} \\
        TAMP~\cite{lee-etal-2025-tamp}  & 54.62 & 48.37 & 1385.71 & 30.00 & 6.73 & 88.40 & 55.78 & 56.36 & 54.04  & 73.75\% \\
        
        \rowcolor{mygreen2}
        ATV-Pruning \scriptsize{(Ours)} & 57.61 & 50.00 & 1474.65 & \textcolor{ForestGreen}{\textbf{31.67}} & \textcolor{ForestGreen}{\textbf{11.13}} & 88.32 & \textcolor{ForestGreen}{\textbf{59.30}} & 55.83 & \textcolor{ForestGreen}{\textbf{55.52}}  & \textcolor{ForestGreen}{\textbf{77.01\%}} \\
        \midrule

        \midrule
        \textcolor{gray}{Qwen2-VL 7B, Dense} & \textcolor{gray}{62.35} & \textcolor{gray}{78.95} & \textcolor{gray}{2306.50} & \textcolor{gray}{50.89} & \textcolor{gray}{52.12} & \textcolor{gray}{88.56} & \textcolor{gray}{84.98} & \textcolor{gray}{82.02} & \textcolor{gray}{68.33} & \multirow{1}*{\textcolor{gray}{100\%}} \\
        \midrule
        \rowcolor{mygray}
        Qwen2-VL 7B & \multicolumn{10}{c}{\textit{60\% Sparsity}}\\
        SparseGPT~\cite{frantar2023sparsegpt} & 56.98 & 71.13 & 1621.23 & 41.00 & \textcolor{ForestGreen}{\textbf{29.50}} & 88.11 & 78.53 & \textcolor{ForestGreen}{\textbf{78.51}} & 62.28 & \multirow{1}*{85.30\%} \\
        Wanda~\cite{sunsimple}  & 54.37 & 69.42 & 1637.66 & 37.22 & 6.30 & 88.38 & 76.85 & 75.76 & 58.84  & \multirow{1}*{77.78\%} \\
        TAMP~\cite{lee-etal-2025-tamp}  & 57.47 & 70.02 & \textcolor{ForestGreen}{\textbf{1923.30}} & 41.22 & 15.93 & 88.40 & 77.49 & 77.24 & 63.65  & 83.79\% \\
        
        \rowcolor{mygreen2}
        ATV-Pruning \scriptsize{(Ours)} & 
        \textcolor{ForestGreen}{\textbf{58.13}} & \textcolor{ForestGreen}{\textbf{71.82}} & 1816.56 & \textcolor{ForestGreen}{\textbf{41.56}} & 23.73 & \textcolor{ForestGreen}{\textbf{88.90}} & 
        \textcolor{ForestGreen}{\textbf{79.08}} & 76.96 & 
        \textcolor{ForestGreen}{\textbf{63.82}}  & \textcolor{ForestGreen}{\textbf{85.65\%}} \\

	\end{tabular}
    }
    \caption{\textbf{Comparison results with state-of-the-art approaches on nine benchmarks under unstructured, uniform pruning.} The official metric for each benchmark and average performance retention (denoted as {Average}) across all benchmarks, are reported to evaluate the performance of pruned models. Best results are highlighted in \textcolor{ForestGreen}{\textbf{green}}}.
    \label{tab:main}
    \vspace{-2em}
\end{table*}

\subsection{Experimental Settings and Details}
\paragraph{Evaluation benchmarks.}
To evaluate our ATV-Pruning, we conduct the experiments on nine standard multimodal benchmarks, including: 
GQA~\cite{hudson2019gqa}, MMBench-EN (MMB)~\cite{liu2024mmbench}, MME (perception and cognition)~\cite{fu2023mme}, MMMU~\cite{yue2024mmmu}, OK-VQA~\cite{marino2019ok}, POPE~\cite{li2023evaluating}, ScienceQA-IMG (SQA$_\text{img}$)~\cite{lu2022learn}, TextVQA~\cite{singh2019towards}, and VizWiz-VQA~\cite{gurari2018vizwiz}. A brief introduction to each benchmark is provided in the Supplementary.
We evaluate the model performance via two metrics: the official metric provided by each benchmark, and \emph{average performance retention} across all benchmarks, computed as the mean value of the pruned model's performance divided by that of its dense counterpart.

\noindent\textbf{Implementation details.}
Our primary experiments are conducted using LLaVA-NeXT (8B)~\cite{liu2024llavanext} at 50\% and 60\% unstructured sparsity levels.
Unless otherwise specified, we fix the global scaling factor at $\alpha=1.0$ when pruning LLaVA-NeXT, which yields on average 15 selected visual tokens per sample per block over the calibration pool.
In addition, we evaluate Qwen2-VL (7B)~\cite{wang2024qwen2} at 60\% sparsity level, and report the experimental results with $\alpha=1.5$.
For fair comparison, we use the same calibration samples across all models and sparsity levels.
Following TAMP~\cite{lee-etal-2025-tamp}, the calibration set contains 128 high-quality image–text pairs randomly sampled from ShareGPT4V~\cite{chen2024sharegpt4v}.
Across all methods, we apply \emph{uniformly distributed unstructured sparsity} to the weight matrices of all linear layers in the Attention and MLP blocks of the LLM backbone.
The compared methods, \emph{i.e.}, SparseGPT~\cite{frantar2023sparsegpt}, Wanda~\cite{sunsimple}, and TAMP~\cite{lee-etal-2025-tamp}, are reproduced using their official configurations.
All experiments run on a single NVIDIA A100 GPU, and we employ lmms-eval~\cite{zhang2025lmms} to ensure a unified evaluation protocol.

\subsection{Comparison with State-of-the-art Approaches}
\label{sec:main-results}
To validate the superiority of our ATV-Pruning, we compare it with the typical pruning approaches for LLMs and LVLMs, including: SparseGPT~\cite{frantar2023sparsegpt} approximates output preservation through local weight updates; Wanda~\cite{sunsimple} incorporates both weights
and activations into the importance evaluation; TAMP~\cite{lee-etal-2025-tamp} performs modality-aware pruning via a multimodal-informed threshold.
Table~\ref{tab:main} reports the performance of the pruned models under unstructured, uniform pruning and summarizes the following findings. 
\begin{itemize}
\item \textbf{\emph{LLaVA-NeXT:}}
At 50\% sparsity, ATV-Pruning achieves the \emph{highest average retention} of \textit{\textbf{94.00\%}}, ranking first on GQA, MME, MMMU, OK-VQA, and VizWiz, while remaining competitive on the other benchmarks. 
At 60\% sparsity, our method again yields the \emph{best average} at \textit{\textbf{77.01\%}}. Despite SparseGPT often being strong at higher sparsity in LLMs with its computationally intensive weight update process~\cite{sunsimple}, ATV-Pruning tops MMMU, OK-VQA, SQA$_\text{img}$, and VizWiz. The advantage over other Wanda-style baselines widens as sparsity increases: at $60\%$, ATV-Pruning exceeds Wanda and TAMP by \textit{\textbf{+12.56}} and \textit{\textbf{+3.26}} average points, respectively. 
Across both pruning levels, ATV-Pruning matches or outperforms these baselines on most tasks.
\item \textbf{\emph{Qwen2-VL:}} 
We further validate the generalizability of our method on Qwen2-VL 7B. At $60\%$ sparsity, ATV-Pruning maintains its lead with the highest average retention of \textit{\textbf{85.65\%}}. It outperforms SparseGPT and TAMP by +0.35 and +1.86 points, respectively, while securing the \textit{\textbf{best results on 6 of 9}} benchmarks, proving its stability..

\end{itemize}
Overall, across both model architectures and pruning levels, ATV-Pruning consistently matches or outperforms state-of-the-art baselines on the majority of tasks, with remarkable pruning efficiency (see Sec.~\ref{sec:pruning_efficiency} for details). We provide additional results for other models and N:M semi-structured sparsity in the Supplementary.

\subsection{Ablation Studies}
\label{sec:ablation}
We conduct controlled ablations to understand each component constituting our ATV-Pruning. Unless specified otherwise, all experiments are performed at 50\% unstructured sparsity level with LLaVA-NeXT (8B) backbone, and evaluated on MMB~\cite{liu2024mmbench}, SQA$_\text{img}$~\cite{lu2022learn}, and VizWiz~\cite{gurari2018vizwiz}. 

\subsubsection{Effect of Scaling Factor \texorpdfstring{$\alpha$}{alpha}}
\begin{figure}[t]
    \centering
    \includegraphics[width=0.95\linewidth]{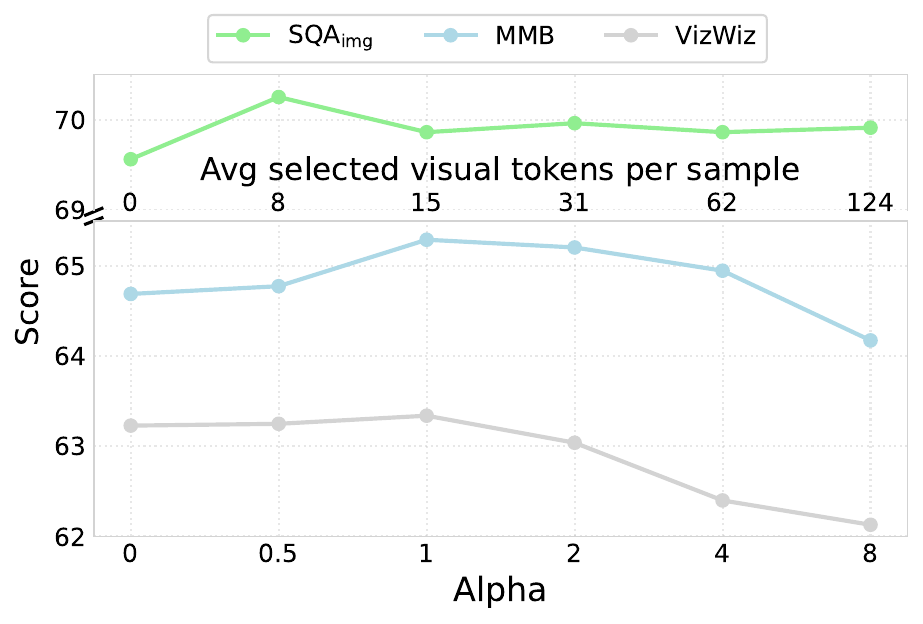}
    \caption{\textbf{Ablation study about global scaling factor $\alpha$.} Larger $\alpha$ allocates more salient visual tokens. A moderate range $\alpha \in [0.5, 2]$ yields stable performance across benchmarks.}
    \label{fig:alpha_tuning}
    \vspace{-1em}
\end{figure}
The global scaling factor $\alpha$ in Eq.~\ref{eq:k_rule_saliency} determines the block-level visual-token budget $K$, which reflects how many salient visual tokens are incorporated into the calibration pool. To validate its effectiveness, we perform the ablation experiments by setting varied $\alpha \in \{0,0.5,1,2,4,8\}$.
Fig.~\ref{fig:alpha_tuning} illustrates that ATV-Pruning achieves competitive results even under the extreme case $\alpha = 0$, where ATV-Pruning holds a text-only calibration pool. For instance,  it reaches 64.69 on MMB, outperforming naive Wanda (63.83)— \emph{highlighting the critical role of text tokens in providing stable activation statistics}. Notably, introducing a moderate amount of visual information with $\alpha \in [0.5,2]$ can improve multimodal robustness, yielding consistently high retention across benchmarks. Thereby, to avoid excessive tuning, we fix $\alpha = 1$ for all LLaVA-NeXT experiments and report results on this setting across all benchmarks in our main experiments.
Beyond this range, \emph{i.e.}, $\alpha \in \{4,8\}$, the performance of ATV-Pruning begins to degrade as redundant visual tokens are incorporated into the calibration pool, which dilutes the linguistic signal and drives the results closer to the modality-agnostic baseline.
The above fact indicates that \emph{maintaining a text-anchored calibration while selectively adding a compact set of high-impact visual tokens strikes superior accuracy.}

\subsubsection{Discussion on Selection Strategy of Visual Tokens}
\label{ablation_selection_strategy}
\renewcommand{\multirowsetup}{\centering}
\definecolor{mygray}{gray}{.92}
\definecolor{mygreen2}{RGB}{238, 243, 243}
\definecolor{ForestGreen}{RGB}{34,139,34}

\begin{table}[t]
    \centering
    
    \footnotesize
    \setlength{\tabcolsep}{3pt}
    \renewcommand{\arraystretch}{1.2}
    \resizebox{0.95\linewidth}{!}{
    \begin{tabular}{l| c | *{3}{>{\centering\arraybackslash}p{1.05cm}} | >{\centering\arraybackslash}p{1.15cm}}
        \textbf{Signal} & \textbf{Adaptive}
        & \textbf{MMB} & \textbf{SQA$_\text{img}$} & \textbf{VizWiz} 
        & \makecell[c]{\textbf{Average}} \\
        \midrule

        \multicolumn{2}{c|}{\textcolor{gray}{Dense}} & \textcolor{gray}{72.16} & \textcolor{gray}{73.28} & \textcolor{gray}{57.65} &  \textcolor{gray}{100\%} \\
        \midrule

        \midrule
        
            Random & \xmark
               & 64.52  & 69.26 & 62.59 & 97.50\% \\
            Drift  & \xmark         
                & 64.95 & 69.56 & 63.30 & 98.24\% \\
        \midrule
            \rowcolor{mygray}
        \multicolumn{6}{c}{\textit{Our ATV-Pruning}} \\
            Drift   & \cmark       
                & \textcolor{ForestGreen}{\textbf{65.29}} & 69.86 & \textcolor{ForestGreen}{\textbf{63.34}} & \textcolor{ForestGreen}{\textbf{98.56\%}} \\

        \midrule

        \midrule
        
            ABS & \cmark
               & 65.12  & \textcolor{ForestGreen}{\textbf{70.65}} & 62.59 & 98.41\% \\
            DBS  & \cmark        
                & 64.78 & 70.10 & 62.84 & 98.14\% \\

    \end{tabular}
    }
    \caption{\textbf{Discussion on selection strategy of visual tokens.} All methods select the same total number of visual tokens for calibration. ``Adaptive'' indicates whether the visual-token budget is dynamically distributed across blocks.
    Our default configuration achieves the best overall retention.}

    \label{tab:visual_selection}
    \vspace{-1em}
\end{table}

In the main experiments, we default to using \emph{visual drift}, \emph{i.e.}, the cosine similarity between a token’s input and output within a block (Eq.~\ref{eq:cosdist}), together with \emph{adaptive visual-token budget allocation} across blocks (Sec.~\ref{sec:adaptive_visual_selection}).
To verify the effectiveness of such setting, we perform the ablation experiments from the following two cases: replacing adaptive saliency-based selection with fixed random selection, denote as ``Random, \xmark''; removing adaptivity and fixing the per-block budget, denote as ``Drift, \xmark'', where all variants are compared under a controlled setting and the \emph{total number of selected visual tokens is kept identical} to ensure fairness.
As shown in Tab.~\ref{tab:visual_selection}, our ATV-Pruning achieves 98.56\% average retention. In comparison, ``Drift, \xmark'' drops to 98.24\%, and the ``Random, \xmark'' drops further to 97.50\%. This implies that (\textit{i}) a meaningful saliency signal (drift) surfaces more informative visual tokens than random selection, and (\textit{ii}) distributing the visual-token budget according to block activity further achieves stable performance.

Additionally, we discuss two alternative visual saliency metrics drawn from recent visual token pruning studies, \emph{i.e.}, \emph{attention-based signal}, denoted as ABS, which measures how much attention each visual token receives from text queries; and \emph{diversity-based strategy}, denoted as DBS, which encourages selecting visually diverse tokens. Implementation details of both are in the Supplementary. 
Both ABS and DBS benefit from adaptive allocation and deliver reasonable results, as reported in Tab.~\ref{tab:visual_selection}.
For instance, ABS achieves the highest SQA$_\text{img}$ score of 70.65. 
However, our ATV-Pruning's adaptive visual drift achieves the best balance across benchmarks and the highest average retention, confirming that per-block representational change serves as an effective and lightweight cue for identifying salient visual tokens during pruning.

\begin{figure}[t]
    \centering
    \includegraphics[width=0.9\linewidth]{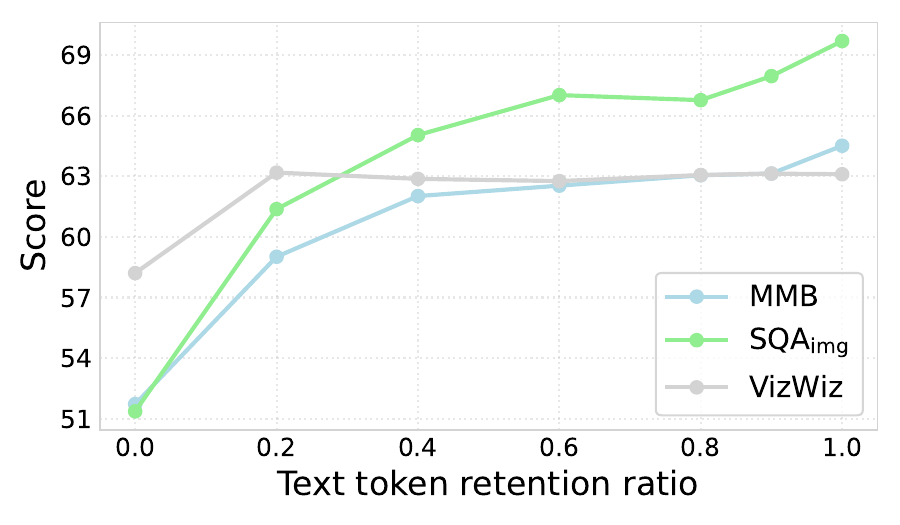}
    \caption{\textbf{Effect of textual token selection under varied retained ratios during calibration}, with the default visual token subset of ATV-Pruning. Removing text tokens consistently harms the performance, highlighting their necessity for stable pruning.}
    \label{fig:text_token_curve}
\end{figure}

\subsection{Further Insights into Textual Token Selection}
\label{sec:text_selection}
Recalling that all textual tokens are included for the calibration pool in Sec.\ref{sec:atv_pruning}, we further discuss whether textual tokens should be adaptively selected during calibration, similar to visual tokens. To investigate this, we progressively reduce the proportion of text tokens used to compute activation norms, with similar drift saliency, while keeping the default visual token selection of ATV-Pruning. We conduct the experiments with varied retained text-token ratio (\emph{i.e.}, \{1.0,0.9,0.8,0.6,0.4,0.2,0\}).
Fig.~\ref{fig:text_token_curve} illustrates that \emph{not selecting all text tokens always leads to performance degradation of ATV-Pruning}. Crucially, for more difficult benchmarks, such as MMB (comprehensive multimodal ability) and SQA$_\text{img}$ (science-oriented reasoning), even reducing the retained text-token ratio from $1.0$ to $0.9$ leads to a noticeable performance drop. For example, MMB decreases from $64.52$ to $63.14$, while SQA$_\text{img}$ drops from $69.71$ to $67.97$, suggesting that even the least-changed text tokens may still contribute to preserving linguistic grounding and cross-modal reasoning. Unlike visual tokens—where selective inclusion improves robustness—the text pathway is not redundant. The above \emph{confirms that full text-token retention for ATV-Pruning provides the most reliable activation statistics for pruning.}

\subsection{Pruning Efficiency Analysis}
\label{sec:pruning_efficiency}

\begin{table}[t]
\centering

\setlength{\tabcolsep}{3pt}
\resizebox{0.9\columnwidth}{!}{%
\begin{tabular}{lccc}
\toprule
\textbf{Method} & \textbf{Time (s)$\downarrow$} & \textbf{Rel. time$\downarrow$} & \textbf{Gain (pp)$\uparrow$} \\
\midrule
Wanda~\cite{sunsimple} & \texttt{73.7} & $1.00\times$ & \texttt{+0} \\
SparseGPT~\cite{frantar2023sparsegpt} & \texttt{666.0} & \texttt{9.04}$\times$ & \texttt{+3.38} \\
TAMP~\cite{lee-etal-2025-tamp} & \texttt{1418.4} & \texttt{19.25}$\times$ & \texttt{+4.31} \\
\rowcolor{mygreen2}
ATV-Pruning (Ours) & \texttt{99.6} & \texttt{1.35}$\times$ & \texttt{+5.64} \\
\bottomrule
\end{tabular}}
\caption{\textbf{Pruning efficiency analysis}, on LLaVA-NeXT (8B) at 50\% sparsity. Time (s): wall-clock pruning time in seconds; Rel. time: pruning time relative to Wanda; Gain (pp): average performance gain in percentage points (pp), taken from Tab.~\ref{tab:main}.}
\label{tab:prune_time}
\vspace{-1em}
\end{table}

To investigate the practical efficiency of ATV-Pruning, we conduct the experiments on LLaVA-NeXT (8B) by measuring the end-to-end wall-clock time to achieve $50\%$ unstructured sparsity under a unified pruning protocol.
The results from Table~\ref{tab:prune_time} suggest that ATV-Pruning (99.6s) exhibits obvious efficiency advantages over SparseGPT (666.0s) and TAMP (1418.4s). The reasons are that SparseGPT introduces a compute-intensive weight update procedure, while TAMP involves additional attention computation and an iterative maximum mean discrepancy (MMD)-based token selection process. 
Notably, compared to Wanda that merely performs a single forward pass to collect activation norms, ATV-Pruning introduces a modest computational overhead (\texttt{1.35}$\times$), while achieving substantial performance advantages (\emph{+5.64} percentage points). In particular, the extra time cost of ATV-Pruning primarily stems from the computation of the cosine-distance-based visual drift $\bar{s}$ in Eq.~\ref{eq:cosdist}. Such fact indicates that ATV-Pruning is capable of serving as a simple, plug-in module for LVLMs pruning.

\section{Conclusion}
In this paper, we introduced ATV-Pruning, a simple yet effective method for pruning LVLMs that explicitly accounts for their asymmetric modal sensitivities. Our approach is based on the core insight that the textual pathway is highly sensitive, while the robust visual pathway possesses significant redundancy. ATV-Pruning leverages this by constructing an asymmetric calibration pool, using all textual tokens but only a sparse, block-adaptively selected subset of visual tokens. Experimental results demonstrate that ATV-Pruning achieves state-of-the-art performance, outperforming prior methods with high pruning efficiency. Future work will focus on extending this asymmetric pruning paradigm to Large Multimodal Models (LMMs), which incorporates a wider range of modalities, including audio, depth, point clouds, etc.

\newpage

\section*{Acknowledgements}
This work was supported in part by Beijing Natural Science
Foundation (L257005), and is also supported by National Natural Science Foundation
of China No. 62441235.

{
    \small
    \bibliographystyle{ieeenat_fullname}
    \bibliography{main}
}

% WARNING: do not forget to delete the supplementary pages from your submission 
\clearpage
\clearpage
\setcounter{page}{1}
\maketitlesupplementary
\appendix
\makeatletter
\renewcommand \thesection{A\@arabic\c@section}
\renewcommand\thetable{A\@arabic\c@table}
\renewcommand \thefigure{A\@arabic\c@figure}
\makeatother

\section{Details of Benchmarks}
\label{sec:benchmarks}
We describe the details of nine benchmarks used in ``\textbf{\emph{Sec.~3.2. Motivation Investigation: Modality-Aware Sensitivity Analysis}}" and ``\textbf{\emph{Sec.~4. Experiments}}", including:

\begin{itemize}
    \item \textbf{GQA}~\cite{hudson2019gqa} evaluates compositional visual reasoning over real-world images. It uses scene graphs from images to generate complex, multi-step questions, aiming to move beyond simple object recognition and reduce the biases found in earlier VQA datasets.
    \item \textbf{MMBench-EN} (referred as \textbf{MMB})~\cite{liu2024mmbench} is the English portion of MMBench, a bilingual multiple-choice benchmark designed for holistic evaluation of large vision–language models. It covers diverse abilities such as perception, reasoning, math, and world knowledge, with carefully curated questions and quality control to enable fair, scalable comparison across models.
    \item \textbf{MME}~\cite{fu2023mme} is a comprehensive evaluation suite for multimodal large language models that explicitly separates perceptual and cognitive skills. It spans 14 subtasks (e.g., existence, OCR, counting, commonsense, code reasoning, etc), using manually designed instruction–answer pairs to reduce data leakage. \emph{We report the sum score of their perception and cognition (all 14 subtasks) evaluation}.
    \item \textbf{MMMU}~\cite{yue2024mmmu} is a challenging benchmark featuring questions from college-level problems across six core disciplines, such as Art \& Design, Science, Health \& Medicine, etc. It is designed to evaluate a model's expert-level knowledge and ability to perform deliberate reasoning with complex, multi-modal information.
    \item \textbf{OK-VQA}~\cite{marino2019ok} is a visual question answering benchmark where answering requires external world knowledge beyond what is directly visible in the image.
    \item \textbf{POPE}~\cite{li2023evaluating} is a targeted benchmark for measuring object hallucination in large vision–language models. It builds yes/no questions about the presence of candidate objects in images to quantify hallucination via accuracy and related metrics.
    \item \textbf{ScienceQA-IMG} (referred as \textbf{SQA$_\text{img}$})~\cite{lu2022learn} is the image-based subset of ScienceQA, a multimodal science question answering benchmark collected from elementary and high school science curricula, emphasizing the chain of thought (CoT) reasoning ability.
    \item \textbf{TextVQA}~\cite{singh2019towards} focuses on answering questions that require reading and understanding text embedded in images.
    \item \textbf{VizWiz}~\cite{gurari2018vizwiz} is a VQA dataset constructed from real visual questions asked by blind and visually impaired users, who capture images with a mobile phone. It features challenging, often low-quality images, making it a realistic benchmark for evaluating practical VQA systems.
\end{itemize}

\renewcommand{\multirowsetup}{\centering}
\definecolor{mygray}{gray}{.92}
\definecolor{mygreen1}{RGB}{253, 244, 244}%255, 234, 234
\definecolor{mygreen2}{RGB}{238, 243, 243}%223, 240, 240
\definecolor{ForestGreen}{RGB}{34,139,34}

\definecolor{Forestred}{RGB}{220,50,50}

\begin{table*}[t]
    \vspace{-0.75em}
    \centering
    \setlength{\tabcolsep}{3pt}
    \footnotesize
    
    \vspace{0.25em}
    \resizebox{\linewidth}{!}{
    \begin{tabular}{l | *{9}{>{\centering\arraybackslash}p{1.13cm}} |>{\centering\arraybackslash}p{1.15cm}}
        % \toprule[1.25pt]
        \textbf{\;Methods} & \textbf{GQA} & \textbf{MMB} & \textbf{MME} & \textbf{MMMU} & \textbf{OKVQA}  & \textbf{POPE} & \textbf{SQA}$_{\text{img}}$ & \textbf{TextVQA} & \textbf{VizWiz}  & \makecell[c]{\textbf{Average}}\\
        
        \midrule
        \textcolor{gray}{LLaVA-NeXT 8B, Dense} & \textcolor{gray}{65.34} & \textcolor{gray}{72.16} & \textcolor{gray}{1965.12} & \textcolor{gray}{40.11} & \textcolor{gray}{60.13} & \textcolor{gray}{87.84} & \textcolor{gray}{73.28} & \textcolor{gray}{65.42} & \textcolor{gray}{57.65} & \multirow{1}*{\textcolor{gray}{100\%}} \\
        % ~ & 100\% & 100\% & 100\% & 100\% & 100\% & 100\% & 100\% & ~ \\
        \midrule
        \rowcolor{mygray}
        LLaVA-NeXT 8B & \multicolumn{10}{c}{\textit{2:4 Semi-Structured Sparsity}}\\
         SparseGPT~\cite{frantar2023sparsegpt} & 55.13 & 42.61 & 1249.34 & 28.89 & 16.79 & \textcolor{ForestGreen}{\textbf{88.21}} & 34.71 & \textcolor{ForestGreen}{\textbf{58.58}} & 53.88  & \multirow{1}*{70.86\%}\\
        Wanda~\cite{sunsimple}  & 54.67 & 37.80 & 1370.97 & 28.00 & 11.07 & 87.52 & \textcolor{ForestGreen}{\textbf{58.08}} & 55.84 & 55.01  & \multirow{1}*{72.62\%} \\
        TAMP~\cite{lee-etal-2025-tamp}  & 56.31 & 42.44 & 1401.91 & 28.67 & 13.84 & 87.46 & 57.76 & 56.11 & 57.15  & 74.90\% \\
        
        \rowcolor{mygreen2}
        ATV-Pruning \scriptsize{(Ours)} & 
        \textcolor{ForestGreen}{\textbf{57.14}} & \textcolor{ForestGreen}{\textbf{46.82}} & \textcolor{ForestGreen}{\textbf{1466.78}} & \textcolor{ForestGreen}{\textbf{29.56}} & \textcolor{ForestGreen}{\textbf{17.91}} & 87.11 & 52.16 & 55.71 & \textcolor{ForestGreen}{\textbf{58.01}}  & \textcolor{ForestGreen}{\textbf{76.29\%}} \\

        \midrule

        \rowcolor{mygray}
        LLaVA-NeXT 8B & \multicolumn{10}{c}{\textit{4:8 Semi-Structured Sparsity}}\\
        SparseGPT~\cite{frantar2023sparsegpt} & 59.72 & 57.39 & 1451.30 & 30.44 & 26.47 & 87.87 & 59.59 & \textcolor{ForestGreen}{\textbf{62.25}} & 58.76 & \multirow{1}*{82.57\%} \\
        Wanda~\cite{sunsimple}  & 60.01 & 59.88 & 1507.97 & 30.67 & 16.60 & \textcolor{ForestGreen}{\textbf{88.33}} & 63.91 & 60.13 & 56.69  & \multirow{1}*{81.52\%} \\
        TAMP~\cite{lee-etal-2025-tamp}  & \textcolor{ForestGreen}{\textbf{60.57}} & 59.19 & 1590.90 & 32.56 & 20.66 & 88.20 & \textcolor{ForestGreen}{\textbf{64.65}} & 60.08 & 57.90  & 83.57\% \\
        
        \rowcolor{mygreen2}
        ATV-Pruning \scriptsize{(Ours)} & 
        60.38 & \textcolor{ForestGreen}{\textbf{60.57}} & \textcolor{ForestGreen}{\textbf{1688.91}} & \textcolor{ForestGreen}{\textbf{32.89}} & \textcolor{ForestGreen}{\textbf{27.23}} & 88.01 & 
        61.68 & 60.44 & 
        \textcolor{ForestGreen}{\textbf{59.70}}  & \textcolor{ForestGreen}{\textbf{85.54\%}} \\

	\end{tabular}
    }
    \vspace{-1em}
    \caption{\textbf{Semi-structured (\emph{2:4; 4:8}) pruning comparisons} on LLaVA-NeXT (8B). Best results are highlighted in \textcolor{ForestGreen}{\textbf{green}}. 
    }
    \label{tab_suppl:semi-structured}
    \vspace{1em}
\end{table*}

\renewcommand{\multirowsetup}{\centering}
\definecolor{mygray}{gray}{.92}
\definecolor{mygreen1}{RGB}{253, 244, 244}%255, 234, 234
\definecolor{mygreen2}{RGB}{238, 243, 243}%223, 240, 240
\definecolor{ForestGreen}{RGB}{34,139,34}
\definecolor{Forestred}{RGB}{220,50,50}

\begin{table*}[t]
    \vspace{-0.75em}
    \centering
    \setlength{\tabcolsep}{3pt}
    \footnotesize
    
    \vspace{0.25em}
    \resizebox{\linewidth}{!}{
    \begin{tabular}{l | *{9}{>{\centering\arraybackslash}p{1.13cm}} |>{\centering\arraybackslash}p{1.15cm}}
        % \toprule[1.25pt]
        \textbf{\;Methods} & \textbf{GQA} & \textbf{MMB} & \textbf{MME} & \textbf{MMMU} & \textbf{OKVQA}  & \textbf{POPE} & \textbf{SQA}$_{\text{img}}$ & \textbf{TextVQA} & \textbf{VizWiz}  & \makecell[c]{\textbf{Average}}\\
        
        \midrule
        \textcolor{gray}{LLaVA-OneVision 7B, Dense} & \textcolor{gray}{62.25} & \textcolor{gray}{80.84} & \textcolor{gray}{1998.62} & \textcolor{gray}{49.22} & \textcolor{gray}{61.00} & \textcolor{gray}{89.13} & \textcolor{gray}{95.93} & \textcolor{gray}{76.05} & \textcolor{gray}{60.38} & \multirow{1}*{\textcolor{gray}{100\%}} \\
        % ~ & 100\% & 100\% & 100\% & 100\% & 100\% & 100\% & 100\% & ~ \\
        \midrule
        \rowcolor{mygray}
        LLaVA-OneVision 7B & \multicolumn{10}{c}{\textit{60\% Sparsity}}\\
         SparseGPT~\cite{frantar2023sparsegpt} & \textcolor{ForestGreen}{\textbf{58.56}} & 73.71 & \textcolor{ForestGreen}{\textbf{1538.61}} & 37.33 & 44.65 & 89.72 & 79.03 & \textcolor{ForestGreen}{\textbf{72.62}} & \textcolor{ForestGreen}{\textbf{62.72}}  & \multirow{1}*{\textcolor{ForestGreen}{\textbf{88.19\%}}}\\
        Wanda~\cite{sunsimple}  & 54.86 & 72.34 & 1293.20 & 37.89 & 25.28 & 89.54 & 78.58 & 69.64 & 61.00  & \multirow{1}*{78.25\%} \\
        TAMP~\cite{lee-etal-2025-tamp}  & 57.64 & 73.20 & 1571.36 & 39.67 & 37.65 & 89.54 & 78.58 & 69.64 & 61.00  & 86.56\% \\
        
        \rowcolor{mygreen2}
        ATV-Pruning \scriptsize{(Ours)} & 
        58.05 & \textcolor{ForestGreen}{\textbf{74.57}} & 1406.75 & \textcolor{ForestGreen}{\textbf{40.11}} & \textcolor{ForestGreen}{\textbf{46.15}} & \textcolor{ForestGreen}{\textbf{89.78}} & \textcolor{ForestGreen}{\textbf{80.47}} & 70.23 & 61.98  & 88.07\% \\

        \midrule

        \midrule
        \textcolor{gray}{Qwen2.5-VL 7B, Dense} & \textcolor{gray}{60.49} & \textcolor{gray}{83.25} & \textcolor{gray}{2333.14} & \textcolor{gray}{50.78} & \textcolor{gray}{42.22} & \textcolor{gray}{87.48} & \textcolor{gray}{-} & \textcolor{gray}{82.93} & \textcolor{gray}{70.48} & \multirow{1}*{\textcolor{gray}{100\%}} \\
        \midrule
        \rowcolor{mygray}
        Qwen2.5-VL 7B & \multicolumn{10}{c}{\textit{60\% Sparsity}}\\
        SparseGPT~\cite{frantar2023sparsegpt} & 54.26 & 72.94 & \textcolor{ForestGreen}{\textbf{2165.54}} & 33.89 & \textcolor{ForestGreen}{\textbf{26.52}} & 88.58 & - & 76.63 & \textcolor{ForestGreen}{\textbf{65.28}} & \multirow{1}*{85.75\%} \\
        Wanda~\cite{sunsimple}  & 51.30 & 73.88 & 1557.82 & 34.78 & 15.80 & \textcolor{ForestGreen}{\textbf{89.16}} & - & 75.60 & 60.53  & \multirow{1}*{78.15\%} \\
        TAMP~\cite{lee-etal-2025-tamp}  & 55.76 & 74.23 & 1974.18 & 39.22 & 22.81 & 88.52 & - & 77.96 & 63.58  & 85.33\% \\
        
        \rowcolor{mygreen2}
        ATV-Pruning \scriptsize{(Ours)} & 
        \textcolor{ForestGreen}{\textbf{57.58}} & \textcolor{ForestGreen}{\textbf{74.91}} & 1977.43 & \textcolor{ForestGreen}{\textbf{39.89}} & 23.57 & 88.56 & 
        - & \textcolor{ForestGreen}{\textbf{78.07}} & 
        64.73  & \textcolor{ForestGreen}{\textbf{86.44\%}} \\

	\end{tabular}
    }\vspace{-1em}

    \caption{\textbf{Additional pruning results for LLaVA-OneVision and Qwen2.5-VL} at 60\% unstructured sparsity, serving as an extension of ``Table 2". Best results are highlighted in \textcolor{ForestGreen}{\textbf{green}}. 
    SQA$_{\text{img}}$ results for Qwen2.5-VL are omitted because the current \texttt{lmms-eval} setup does not yield reliable evaluations.}
    \label{tab_suppl:more_models}
\end{table*}

\section{Beyond Unstructured Pruning: Additional Discussion on Semi-Structured Pruning}

While the main paper focuses on the flexible unstructured pruning to demonstrate the fundamental capability of ATV-Pruning, here we complement those results by evaluating our approach under \emph{\textbf{hardware-friendly semi-structured} sparsity in the general \textbf{N:M} setting}. This format imposes fixed, non-negotiable structural constraints crucial for enabling efficient inference acceleration on modern GPUs. In the $N{:}M$ setting, in each group of $M$ consecutive weights exactly $N$ remain non-zero. Specifically, we study the NVIDIA-compatible 2:4 pattern and the more flexible 4:8 variant, both yielding uniformly distributed 50\% sparsity across linear layers. The 2:4 format, in particular, is natively supported by NVIDIA Ampere and Hopper architectures and allows direct use of sparse Tensor Cores~\cite{mishra2021accelerating}; practical acceleration for LLM backbones (e.g., LLaMA~\cite{touvron2023llama}) commonly falls in the range of $1.2\times$–$1.6\times$~\cite{sunsimple}.

Tab.~\ref{tab_suppl:semi-structured} reports results on LLaVA-NeXT (8B)~\cite{liu2024llavanext}, where we apply $\alpha=1.0$ for ATV-Pruning in both patterns. As expected, semi-structured pruning is more restrictive than unstructured sparsification, causing all methods to experience noticeable degradation relative to the dense model. Nonetheless, ATV-Pruning consistently shows the strongest robustness: it achieves the highest average retained performance of \emph{\textbf{76.29\%}} under 2:4 and \emph{\textbf{85.54\%}} under 4:8, outperforming all baselines by a clear margin and leading on most benchmarks. Notably, the gains persist across both the stricter 2:4 constraint and the more relaxed 4:8 setting, indicating that our pruning criterion reliably preserves essential weights even under rigid structural patterns. These results highlight ATV-Pruning as a practical and effective solution for accelerating LLMs on widely used hardware backends.

\section{An Extension of ``\emph{Tab.~2}": Additional Results on Other Backbone Models}
To further examine the robustness and generality of our approach, we extend ATV-Pruning to two additional LVLMs: LLaVA-OneVision (7B)~\cite{li2024llava} and Qwen2.5-VL (7B)~\cite{bai2025qwen25}, with $\alpha=1.0$. The comparison results at 60\% unstructured sparsity are reported in Table~\ref{tab_suppl:more_models}, which serves as \emph{\textbf{an extension of ``Tab.~2"}}. Overall, ATV-Pruning continues to deliver strong performance across architectures, confirming its effectiveness beyond the models discussed in the main paper.

Across both models, ATV-Pruning consistently outperforms Wanda-style baselines (Wanda~\cite{sunsimple} and TAMP~\cite{lee-etal-2025-tamp}) by a clear margin. For instance, on LLaVA-OneVision, it improves average retention by \textit{\textbf{+9.82}} and \textit{\textbf{+1.51}} points over Wanda and TAMP, respectively; on Qwen2.5-VL, the gains are \textit{\textbf{+8.29}} and \textit{\textbf{+1.11}}. This underscores that ATV-Pruning better preserves multimodal activation statistics during calibration compared to other activation-aware methods.

Notably, ATV-Pruning is also \emph{competitive with, and often superior to,} SparseGPT~\cite{frantar2023sparsegpt}. It delivers the best results on a greater number of benchmarks across both models, closely tracking SparseGPT’s average performance on LLaVA-OneVision (88.07\% vs.\ 88.19\%) while achieving the highest overall average on Qwen2.5-VL with \emph{\textbf{86.44\%}}—a \emph{\textbf{+0.69}} point improvement over SparseGPT.

In summary, across all four tested models (LLaVA-NeXT, Qwen2-VL, LLaVA-OneVision, and Qwen2.5-VL) at high sparsity (60\%), ATV-Pruning consistently exhibits superior stability, leading on more individual benchmarks than other methods. Given our high performance retention at a pruning cost only marginally higher than Wanda (as shown in  ``\textbf{\emph{Sec.~4.5. Pruning Efficiency Analysis}}"), ATV-Pruning provides the optimal trade-off between performance and efficiency.

\section{Experimental Details  for ``\emph{Sec.~3.2.  Motivation Investigation: Modality-Aware Sensitivity Analysis}"}
We detail the experimental setup for the motivation investigation in Sec.~3.2, involving the following aspects:

\noindent \textbf{Adopted Benchmarks.}
Our sensitivity analysis is conducted on three standard LVLM benchmarks: MMBench-EN (MMB)~\cite{liu2024mmbench}, ScienceQA-IMG (SQA\textsubscript{img})~\cite{lu2022learn}, and VizWiz~\cite{gurari2018vizwiz}. Consistent with our main experiments, all evaluations are managed via the \texttt{lmms-eval} toolkit~\cite{zhang2025lmms} and adhere to the official metrics for each benchmark.

\noindent \textbf{Implementation Details.}
All analyses use LLaVA-NeXT (8B)~\cite{liu2024llavanext} as the backbone model. Activation statistics are estimated from a calibration pool of 128 high-quality image-text pairs from ShareGPT4V~\cite{chen2024sharegpt4v}, identical to the calibration set used in our main experiments. The Mixture-of-Transformer (MoT)~\cite{shi2024lmfusion, deng2025emerging} analysis probe is constructed by replicating the QKV and FFN linear layers within each Transformer block to create distinct textual and visual pathways. A modality mask routes tokens to their respective pathway. Other components, such as residual connections, positional encodings, and the core attention logic, remain shared and identical. Pruning is then applied independently to each pathway under the three calibration configurations: text-only, image-only, and mixed.

\section{Implementation Details for ``\emph{Sec.~4.3.2 Discussion on Selection Strategy of Visual Tokens}"}
Recalling Sec.~4.3.2, we evaluate two alternative saliency signals for visual token selection: \emph{(i)} an \textbf{attention-based signal (ABS)} and \emph{(ii)} a \textbf{diversity-based signal (DBS)}. These serve as ablative comparisons to our proposed visual drift metric (defined in Eq.~4). Here, we further provide their corresponding implementation details.

\subsection{Attention-based Signal (ABS)}
For the ABS strategy, we replace the visual drift saliency score in Eq.~4. The new saliency $s_v$ for a visual token $v$ at block $b$ is defined as the average cross-attention score it receives from all text tokens $\mathcal{T}$ in the prompt.

Let $A^{(b)}_{t \to v}$ be the attention weight from a text token $t \in \mathcal{T}$ to a visual token $v$ in the cross-attention mechanism of block $b$. The saliency score is:
\[
s_v = \frac{1}{|\mathcal{T}|} \sum_{t \in \mathcal{T}} A^{(b)}_{t \to v}
\]
A higher $s_v$ indicates that the visual token receives more attention from the text prompt, implying greater saliency. This attention-based $s_v$ then replaces the visual drift score in all subsequent calculations: it is used to calculate the block-average saliency $\bar{s}$ (Eq.~5), determine the block-adaptive budget $K$ (Eq.~6), and perform the final $\mathrm{TopK}$ selection (Eq.~7). This approach is inspired by prior attention-based visual token pruning work~\cite{chen2024image, zhao2025stitch}.

\subsection{Diversity-based Signal (DBS)}
The DBS strategy involves a two-part modification, replacing both the saliency definition (Eq.~4) and the selection mechanism (Eq.~7).

First, to maintain the block-adaptive budgeting framework, we define a per-token \emph{diversity score} as the saliency $s_v$ in Eq.~4. This score measures the average cosine distance of a visual token $v$ from all other visual tokens $\mathcal{V}$ in the same sample, using their input representations $\mathbf{X}_{\mathrm{in}}$:
\[
s_v = \frac{1}{|\mathcal{V}| - 1} \sum_{u \in \mathcal{V}, u \neq v} \left( 1 - \cos(\mathbf{X}_{\mathrm{in},v}, \mathbf{X}_{\mathrm{in},u}) \right)
\]
This score $s_v$ is used \emph{only} to set the block-adaptive budget. Specifically, it is aggregated across all visual tokens to compute the block-average saliency $\bar{s}$ (Eq.~5), which in turn determines the per-sample budget $K$ (Eq.~6).

Second, we replace the simple $\mathrm{TopK}$ selection (Eq.~7). Note that we do not simply take the $\mathrm{TopK}$ of these $s_v$ scores. A high $s_v$ only indicates high average diversity. It is possible for the $K$ tokens with the highest $s_v$ to be clustered together (e.g., all are distant from the majority of tokens, but form a small, tight cluster themselves). Therefore, to ensure the selected subset is \emph{internally diverse}, we replace the $\mathrm{TopK}$ rule with a \textit{max-min diversity selection scheme}~\cite{alvar2025divprune}. This scheme directly selects the subset $\mathcal{V}_{\text{sub}}$ of size $K$ that maximizes the minimum pairwise distance among its members, operating directly on the token representations. Formally, the selection rule (replacing Eq.~7) becomes:
\[
\mathcal{V}_{\text{sub}} = \underset{\mathcal{S} \subset \mathcal{V}, |\mathcal{S}|=K}{\operatorname{argmax}} \left( \min_{u, w \in \mathcal{S}, u \neq w} d(\mathbf{X}_{\mathrm{in},u}, \mathbf{X}_{\mathrm{in},w}) \right)
\]
where $d(\cdot, \cdot)$ is the cosine distance $1 - \cos(\cdot, \cdot)$. This directly optimizes for a diverse subset, avoiding the issue where $\mathrm{TopK}$ selection might choose $K$ tokens that are individually diverse on average but are all clustered in a similar region of the representation space, lacking internal diversity.

\section{An Extension of ``\emph{Sec.~4.3.2 Discussion on Selection Strategy of Visual Tokens}": Additional Visualization of Visual Drift across Blocks}
\begin{figure}[t]
    \centering
    \includegraphics[width=1\linewidth]{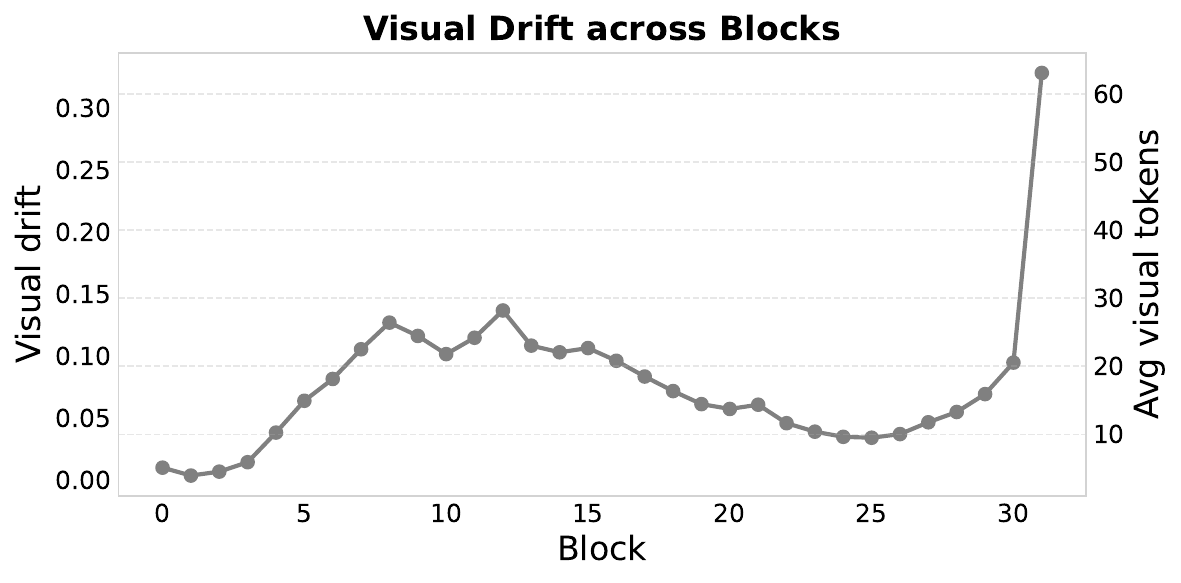}
    \caption{\textbf{Trend of ATV-Pruning's block-wise Visual Drift and Adaptive Token Allocation} on LLaVA-NeXT (8B) with $\alpha=1.0$. The left y-axis measures the average visual drift per block. The right y-axis indicates the corresponding average number of visual tokens selected per sample. The trend illustrates the block-adaptive nature of our method: more tokens are retained in layers where visual representations undergo significant updates.}
    \label{fig:visual_drift}
    \vspace{-1em}
\end{figure}

To better understand the behavior of our proposed Block-Adaptive Visual Selection (Sec.~3.3.2), we further visualize the block-wise statistics of visual drift. Recall that we define visual saliency based on the representation drift, quantified as the cosine distance between the input and output features of a visual token within a block (Eq.~4). Consequently, the budget of visual tokens retained, $K$, is directly proportional to the average saliency $\bar{s}$ of that block (Eq.~5). 

Fig.~\ref{fig:visual_drift} illustrates the progression of visual drift across the 32 Transformer blocks of LLaVA-NeXT (8B)~\cite{liu2024llavanext} with $\alpha=1.0$ (default setting). We observe that visual drift is highly non-uniform across the blocks. Specifically, drift is relatively low in the early stages (Blocks 0–3), indicating stable feature representations. In contrast, the trend highlights two distinct zones of high activity: a broad elevation in the middle layers (peaking around Blocks 8–12) and a sharp, significant spike in the final layer (Block 31).
By explicitly coupling the visual token budget $K$ to this trend, ATV-Pruning acts dynamically: it aggressively minimizes visual tokens in ``visually stable'' blocks to allow essential text tokens to dominate the calibration pool, while automatically preserving a higher density of visual tokens for calibration in blocks where significant visual feature transformations occur.
% \clearpage
% {
%     \small
%     \bibliographystyle{ieeenat_fullname}
%     \bibliography{supplementary}
% }

\end{document}